\documentclass[runningheads,envcountsame,a4paper]{llncs}
\usepackage[T1]{fontenc}
\usepackage{lmodern}
\usepackage{textcomp}
\usepackage[pdftex]{graphicx}
\usepackage{multirow}
\usepackage{tabularx}
\usepackage{epstopdf}
\usepackage{microtype}
\begin{document}
\newcolumntype{L}[1]{>{\raggedright\arraybackslash}p{#1}}
\newcolumntype{C}[1]{>{\centering\arraybackslash}p{#1}}
\newcolumntype{R}[1]{>{\raggedleft\arraybackslash}p{#1}}
\title{Investigation on N-gram Approximated RNNLMs for Recognition of\\
Morphologically Rich Speech
%\thanks{The research was partly supported by the DANSPLAT (EUREKA\_15\_1\_2016-0019) project.}
}
\titlerunning{N-gram Approximated RNNLMs for Morphologically Rich ASR}
% If the paper title is too long for the running head, you can set
% an abbreviated paper title here
%
\author{Bal\'azs Tarj\'an\inst{1,2} 
\and
Gy\"orgy Szasz\'ak\inst{1}
\and
Tibor Fegy\'o\inst{1,2}
\and
P\'eter Mihajlik\inst{1,3}
}
\authorrunning{B. Tarj\'an et al.}
% First names are abbreviated in the running head.
% If there are more than two authors, 'et al.' is used.
%
\institute{Department of Telecommunications and Media Informatics, \\
Budapest University of Technology and Economics, Budapest, Hungary \\
\email{$\{$tarjanb,szaszak$\}$@tmit.bme.hu}
\and
SpeechTex Ltd., Budapest, Hungary \\
\email{fegyo@speechtex.com}
\and
THINKTech Research Center, Budapest, Hungary \\
\email{mihajlik@thinktech.hu}
}

\maketitle              % typeset the header of the contribution
\setcounter{footnote}{0}

\begin{abstract}
Recognition of Hungarian conversational telephone speech is challenging due to the informal style and morphological richness of the language. Recurrent Neural Network Language Model (RNNLM) can provide remedy for the high perplexity of the task; however, two-pass decoding introduces a considerable processing delay. In order to eliminate this delay we investigate approaches aiming at the complexity reduction of RNNLM, while preserving its accuracy. We compare the performance of conventional back-off n-gram language models (BNLM), BNLM approximation of RNNLMs (RNN-BNLM) and RNN n-grams in terms of perplexity and word error rate (WER). Morphological richness is often addressed by using statistically derived subwords - morphs - in the language models, hence our investigations are extended to morph-based models, as well. We found that using RNN-BNLMs 40\% of the RNNLM perplexity reduction can be recovered, which is roughly equal to the performance of a RNN 4-gram model. Combining morph-based modeling and approximation of RNNLM, we were able to achieve 8\% relative WER reduction and preserve real-time operation of our conversational telephone speech recognition system.

\end{abstract}

\section{Introduction}

Recognition of conversational telephone speech poses great challenge due to the low acoustic quality (limited bandwidth, speaker noises, lossy compression etc.) on the one hand and high perplexity of spontaneous speaking style on the other hand. The less constrained grammar and word order of informal speech make the language model estimation less accurate due to the increased variability of both the individual words and their possible sequential combinations. Data sparsity issues caused by morphological richness of the language or lack of sufficient training data make the problem even harder.

In the last few years neural networks have been successfully applied in the field of language modeling~\cite{Arisoy2014,mikolov2010recurrent}. Recurrent networks have proved particularly efficient for the task~\cite{mikolov2010recurrent} especially if they exploit Long Short-Term Memory (LSTM) units~\cite{Hochreiter1997,sundermeyer2012lstm}. However, the RNNLMs have a vast amount of inner states that makes their usage in the first-pass of an Automatic Speech Recognition (ASR) system computationally infeasible. RNNLMs hence are usually utilized in a second decoding pass for rescoring the hypotheses obtained with a less heavy LM. The two-pass decoding, however, introduces a considerable processing delay~\cite{Enarvi2017,sundermeyer2012lstm}.

Various techniques have been proposed to address direct applicability of RNNLMs in the single-pass decoding scheme. A possible solution is to approximate the probability distributions of RNNLMs with conventional back-off n-gram language models~\cite{Adel2014,Arisoy2014,Deoras2011}. Although the converted model (RNN-BNLM) loses its ability to model long contexts and distributed input features, it can be directly applied for first-pass decoding that makes these techniques attractive. Recently another approach called RNN n-gram has also been introduced~\cite{Chelba2017}. RNN n-gram language models are special recurrent networks trained on n-grams sampled from the training data. As a consequence, the size of the modeled context here is also limited, but RNN n-gram models are able to learn word embeddings just like standard RNNLMs.

Our ambition in this paper is to compare conventional BNLMs, RNNLMs and n-gram approximated RNNLMs in a morphologically very rich language, Hungarian. The rich morphology of Hungarian allows for a weakly constrained word order, and per se, results in an extreme large vocabulary. We, moreover, go for spontaneous speech. All of these three effects -- varying word order, large vocabulary and spontaneity -- hamper statistic models' ability to yield consistent estimates by high confidence. Since data sparsity issues can be often handled by estimating language models on statically derived subword units (morphs)~\cite{Creutz2002,Kurimo2006487,Mihajlik2010} in morphologically rich languages, we extended our investigation to morph-based language models, as well.

Besides the related work already cited, another paper, written by T\"uske et al.~\cite{Tuske2018} is also closely related to our work. In this comprehensive study a RNNLM, RNN n-gram models and BNLMs are compared on various English and German ASR tasks. RNN n-gram models were found to be superior to BNLMs both in terms of word perplexity and WER, whereas high order RNN n-grams were close to the performance of an unrestricted RNNLM. However, in~\cite{Tuske2018} ASR results were obtained with two-pass decoding, and German (and obviously English) morphology is less complex than Hungarian.

Although subword language modeling has been used in morphologically rich Finnish ASR systems for more than a decade now~\cite{Creutz2002,Kurimo2006487}, it was not found beneficial for spontaneous conversational speech until recently. In~\cite{Enarvi2017}, subword RNNLMs were trained on Finnish and Estonian conversations and used for rescoring lattices generated with conventional back-off models. Subword language models have already been applied successfully for recognition of Hungarian conversational speech~\cite{Mihajlik2010,Tarjan2013b}, but neural language models have not been used before to the best of our knowledge. We found only one mention of application of morph-based approximated RNNLMs in the first pass of an ASR system~\cite{Singh2018}, however this paper did not provide morph-based ASR results.

Overall, we consider the main contributions of our work are (1) presenting the first ASR results with using morph-based RNN-BNLMs in single-pass decoding; (2) comparing the performance of BNLMs and n-gram approximations of RNNLM (RNN n-gram models, RNN-BNLMs); (3) carrying out for the first time an evaluation of neural language models on very rich morphology Hungarian for speech recognition tasks on spontaneous speech; and (4) doing this preserving real-time operation capabilities by low delay.

In next section the experimental database is introduced along with the applied preprocessing techniques. In Section~\ref{lm} we describe the techniques we used for training our different types of language models. Next Section~\ref{results} presents the experimental results, while in the conclusions we highlight the most impactful outcome of our work.

\section{Database} \label{database}

\subsection{Training Data}

\subsubsection{Original Data}

Our experiments were performed on anonymised manual transcripts of telephone customer service calls which were collected from the Hungarian Call Center Speech Database (HCCSD). We selected 290 hours of recordings from HCCSD for training purposes. The corresponding transcripts that were used for building the language models consisted of 3.4 million word tokens and contained 100,000 unique word forms. In order to accelerate the training of recurrent networks only the most frequent 50,000 word forms were retained in the final vocabulary. Out-Of-Vocabulary (OOV) words and sentence endings were replaced with $\langle unk\rangle$ and $\langle eos\rangle$ symbols respectively.

\subsubsection{Morph Segmented Data}

Language modeling of morphologically rich languages poses a great challenge, since the large number of word forms cause data sparseness and high OOV rate. A common remedy is to segment words into smaller parts and train language models on these subword sequences~\cite{Kurimo2006487,Mihajlik2010}. One of the most popular statistical word segmentation algorithm is Morfessor~\cite{Creutz2002}, which was specifically designed for processing morphologically rich languages. We applied the Python implementation of the original algorithm called Morfessor 2.0~\cite{Virpioja2013a}. Hyperparameters of the segmentation were optimized on the validation test set (see Section \ref{test_data}).

Morph segmentation increased the number of tokens in the training text with around 12\% (from 3.4 million to 3.8 million). However, number of types decreased to around 32,000 from 100,000. In order to provide sufficient amount of training samples to $\langle unk\rangle$, the vocabulary size of morph-based models was limited in 30,000 morphs. Sentence endings were replaced with $\langle eos\rangle$ just like in the case of word-based text data. Non-initial morphs of every word were tagged to provide information to the ASR decoder for the reconstruction of word boundaries (see left-marked style in~\cite{Smit}).

\subsection{Test Data} \label{test_data}

Almost 20 hours of conversations were selected from HCCSD for testing purposes. The test dataset was split into two disjoint parts (see Table~\ref{test_stat}). The validation set ({\raise.17ex\hbox{$\scriptstyle\sim$}}7.5 hours) and the corresponding text transcripts were used for optimization of the hyperparameters (e.g. learning rate control, early stopping), whereas evaluation set ({\raise.17ex\hbox{$\scriptstyle\sim$}}12 hours) was used to test the models and report experimental results. Morph-based segmentation of evaluation dataset was performed with Morfessor 2.0 toolkit using the segmentation model we optimized on the validation set.

\begin{table}[htbp]
\caption{Test database statistics}
\begin{center}
\begin{tabular}{R{3cm}C{2.5cm}C{2.5cm}}
\hline
                   & \textbf{Validation} & \textbf{Evaluation} \\ \hline
Duration {[}h:m{]} & 7:31                & 12:12               \\
\# of word tokens       & 45773               & 66312               \\
\# of morph tokens       & 57849               & 84385               \\
word OOV rate {[}\%{]}  & 2.7                 & 2.5                 \\
morph OOV rate {[}\%{]}  & 0.07                 & 0.08                 \\ \hline
\end{tabular}
\label{test_stat}
\end{center}
\end{table}

\section{Language Modeling} \label{lm}

\subsection{Back-off N-gram Models}

The conventional, count-based, back-off language models (BNLMs) were trained using the SRI language modeling toolkit~\cite{Stolcke2002}. In order to maximize their performance, the baseline BNLMs applied neither count-based n-gram cut-offs nor entropy-based pruning~\cite{Stolcke2000}. All BNLMs were estimated on cross-sentence n-grams and smoothed with Chen and Goodman's modified Kneser-Ney discounting~\cite{Chen1999}.

\subsection{Recurrent Language Model} \label{rnnlm}

\begin{figure}[htbp]
\centerline{\includegraphics[width=0.75\textwidth]{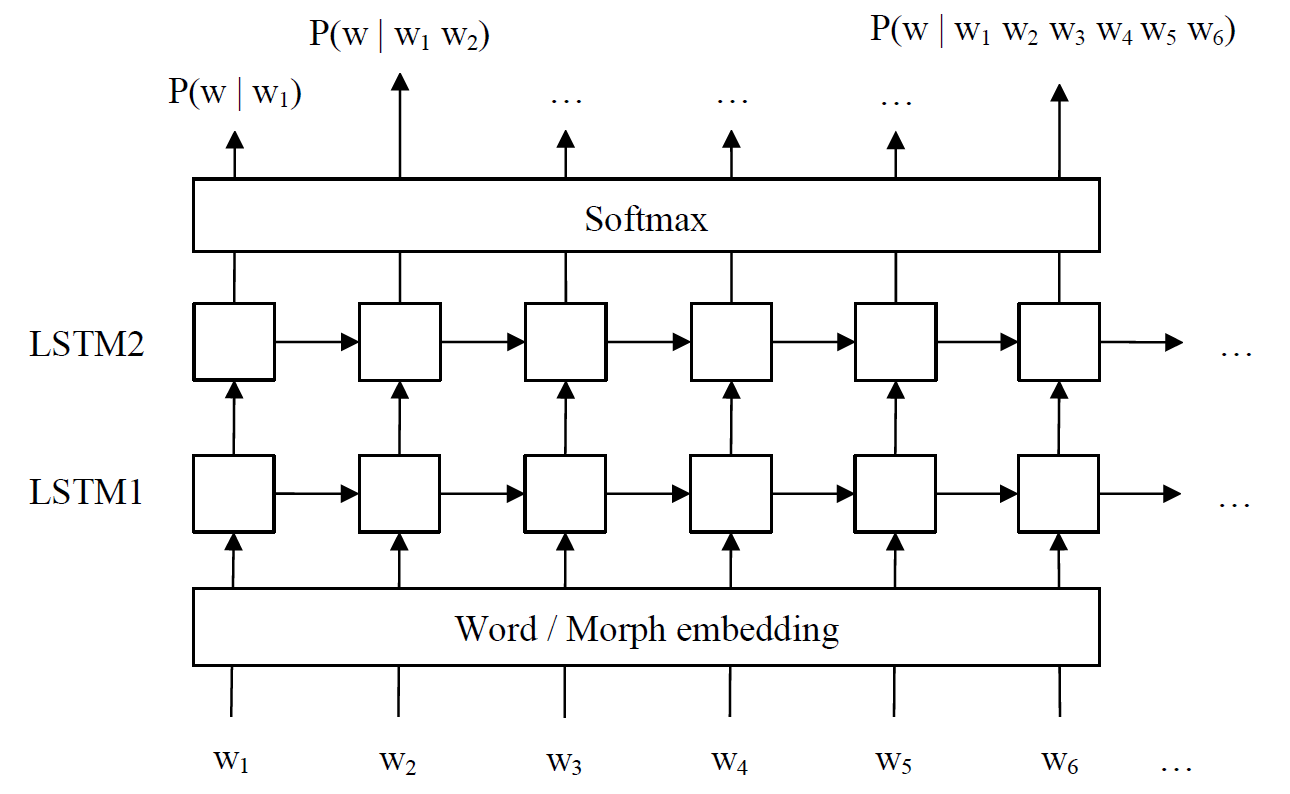}}
\caption{The recurrent LSTM language model structure used in our experiments}
\label{fig:stateful}
\end{figure}

The 2-layered LSTM RNNLM structure we used in our experiments is illustrated in Fig.~\ref{fig:stateful}. This type of network has already been successfully applied for other language modeling tasks~\cite{Chelba2017,Zaremba2014}. Our implementation\footnote{https://github.com/btarjan/stateful-LSTM-LM} is based on the TensorFlow sample code of the Penn Tree Bank language model presented in~\cite{Zaremba2014}.

The hyperparameters of the neural network were optimized on the validation set. One batch consists of 32 sequences containing 35 tokens each (words or morphs). LSTM states are preserved between the batches, so \textit{stateful} recurrent networks are trained according to TensorFlow terminology. The 650 dimension word/morph embedding vectors are trained on the input data, since we did not find any benefit of Hungarian pretrained embeddings. In order to match the dimensionality of embeddings the output dimension of LSTM neurons is also set to 650. After testing several optimization algorithms, we decided on the momentum accelerated, Stochastic Gradient Descent (SGD). The initial learning rate was set to 1, which is halved after every epoch where the cross entropy loss increases. For regularization purposes, dropout layers with keep probability of 0.5 and early stopping with patience of 3 epochs are applied. 

\subsection{RNN N-grams}

Although RNNLM can model word sequences with outstanding accuracy~\cite{mikolov2010recurrent,sundermeyer2012lstm}, the need for large context prevents its practical use in many cases. The modeled context can be reduced if we organize training data into n-grams~\cite{Chelba2017}. It was shown that this limitation of history length does not necessarily have a drastic impact on perplexity~\cite{Tuske2018}. 

\begin{figure}[htbp]
\centerline{\includegraphics[width=0.75\textwidth]{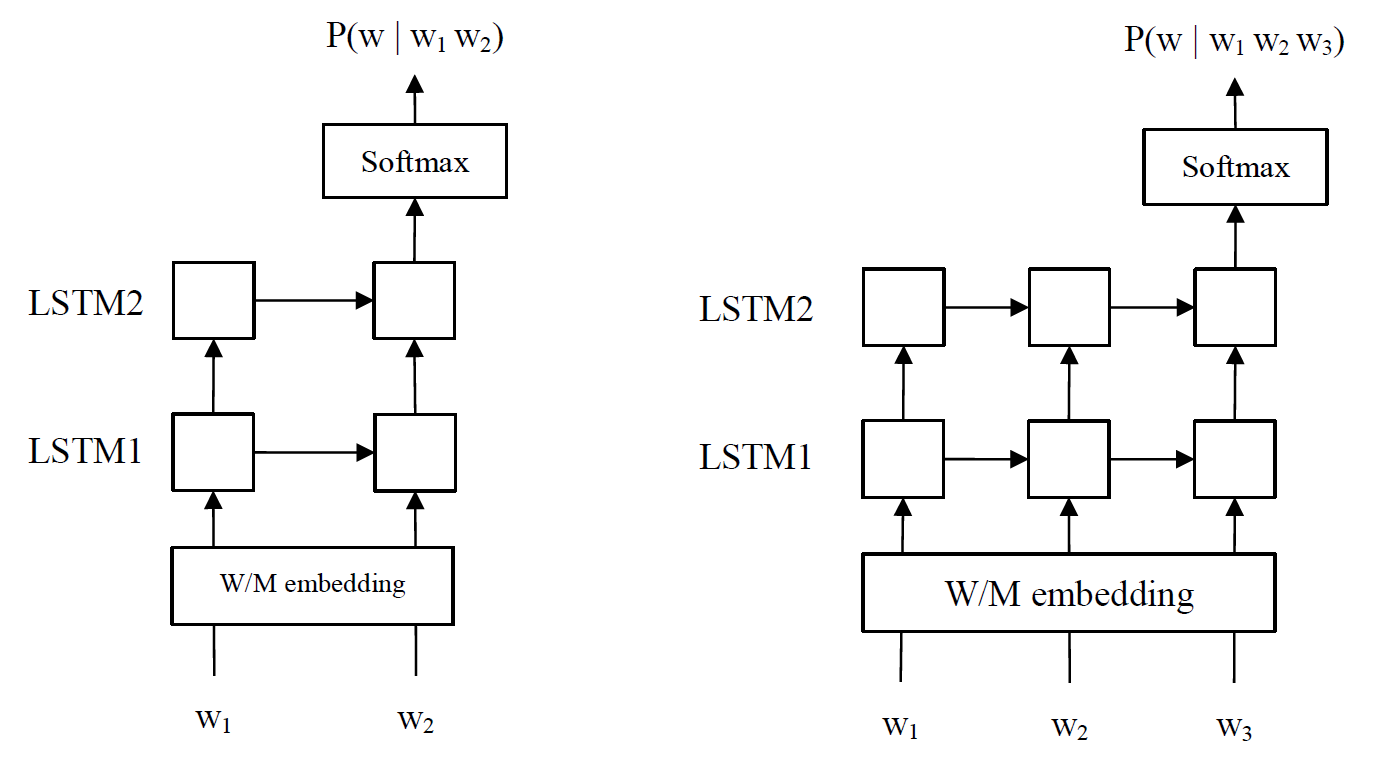}}
\caption{Two examples for the applied RNN n-gram structures (3-gram and 4-gram)}
\label{fig:lstm_ngram}
\end{figure}

Two examples for the many-to-one structure of our RNN n-gram implementations are illustrated in Fig.~\ref{fig:lstm_ngram}. The hyperparameters and optimization used in RNN n-gram training were the same as those we applied for the RNNLM -- except for the sequence length and batch size. Sequence length of RNN n-grams depends on the actual n-gram order (n-1), whereas -- thanks to the shorter sequences -- RNN n-grams can use a larger batch size (512). An additional important difference compared to RNNLM is that RNN n-gram models do not apply dropout between the two LSTM layers. 

\subsection{Approximation of RNNLM with BNLM}

There are various approaches for the approximation of an RNNLM with a back-off ngram language model~\cite{Adel2014,Deoras2011}. In~\cite{Adel2014} three methods are compared and a text generation based approximation is suggested. The main idea of this approach is that the BNLM is estimated from a large text which was generated with the RNNLM. For training the RNN-BNLM models we generated 100 million words/morphs with the corresponding RNNLM (RNN-BNLM 100M). In order to assess the importance of corpus size, we generated a text with 1 billion morphs (RNN-BNLM 1B), as well. The generation of 1 billion morphs took around one week with four NVIDIA GTX 1080 Ti GPUs. Note, that perplexity results in section~\ref{ppl_results} were measured with unpruned RNN-BNLM models, whereas RNN-BNLMs used in ASR decoding are pruned to limit runtime memory usage.

\section{Experimental Results} \label{results}

In the first part of this section, we present perplexities measured on the evaluation text set of our conversational speech database. We compare the performance of the language modeling techniques that were described in Section~\ref{lm}. Our aim is to measure the perplexity reduction that can be achieved with RNNLM compared to BNLMs and how much of this reduction can be preserved with the n-gram approximated models. In the second part, we utilize these language models in an ASR system to show whether the application of subwords and approximated RNNLMs can turn to reduction in WER.

\subsection{Perplexity Results} \label{ppl_results}

All perplexity results were measured with cross-sentence language models as it was discussed in section~\ref{lm}. Note that BNLMs and RNN-BNLMs were estimated only up to 6-grams as larger model order did not result significant reduction in perplexity.

\subsubsection{Word-based Models}
Perplexity results of word-based models are shown in Table~\ref{word_ppl}. What can be clearly seen at the first glance is the superiority of RNN-based language models over BNLMs. Perplexity of RNN n-gram models improves step by step as we increase the modeling context, while results of BNLMs saturates at around 5-gram. This can be explained by the fact that recurrent models can provide a more accurate probability estimate for unobserved n-grams with the help of distributed modeling of input tokens.

\begin{table}[t]
\centering
\caption{Perplexities of word-based backoff n-gram, RNN n-gram and backoff approximated RNN language models as function of n-gram order}
\begin{tabular}{C{1cm}C{1.5cm}C{2.25cm}C{2.25cm}C{1.5cm}}
\hline
\textbf{order} & \textbf{BNLM} & \textbf{\begin{tabular}[c]{@{}c@{}}RNN-BNLM\\ 100M\end{tabular}} & \textbf{\begin{tabular}[c]{@{}c@{}}BNLM + \\ RNN-BNLM\\ 100M\end{tabular}} & \textbf{\begin{tabular}[c]{@{}c@{}}RNN\\ n-gram\end{tabular}} \\ \hline
2              & 130.8         & 136.3                                                            & 125                                                                        & 124.4                                                         \\
3              & 92.4          & 94.5                                                             & 82.5                                                                       & 77.8                                                          \\
4              & 85.7          & 86.8                                                             & 74.4                                                                       & 64.2                                                          \\
5              & 84.4          & 85.5                                                             & 72.8                                                                       & 58.3                                                          \\
6              & 84.1          & 85.4                                                             & 72.5                                                                       & 54.9                                                          \\
8              &               &                                                                  &                                                                            & 52.4                                                          \\
10             &               &                                                                  &                                                                            & 49.5                                                          \\
14             &               &                                                                  &                                                                            & 47.1                                                          \\
18             &               &                                                                  &                                                                            & 46.4                                                          \\ \hline
inf.           &               &                                                                  &                                                                            & 44.6                                                          \\ \hline
\end{tabular}
\label{word_ppl}
\end{table}

In the last row of Table~\ref{word_ppl}, where the order of context is indicated with infinite (inf.), we can find the perplexity of the LSTM RNN language model (see section~\ref{rnnlm}). This implies that this model takes (theoretically) all previous words into account to estimate probability. RNNLM can halve the perplexity of conventional BNLM, however as RNN n-gram results suggest it is only partly due to the modeling of large context, but also due to the previously mentioned generalization abilities of RNNs~\cite{Tuske2018}.

The perplexity of the BNLM approximation of the RNNLM (RNN-BNLM 100M) is slightly worse, but very close to the perplexity of the original BNLM. The interpolated model (BNLM + RNN-BNLM 100M), however, improves perplexity with around 10-15\% which suggests that there are different n-gram probability distributions behind the similar perplexities. If we would like to capture the effectiveness of RNNLM approximation, we could say that the performance of a pure BNLM model is bit worse than a RNN 3-gram, while the interpolated language model is slightly better than the RNN 3-gram.

We can get an even better insight to the benefit of n-gram approximated RNNLMs, if we estimate the perplexity reduction associated with each approach. Assuming that we utilize 4-gram language models which usually represent a good trade-off between precision and memory consumption, the total amount of perplexity improvement between the baseline 4-gram BNLM (85.7) and the LSTM RNNLM (44.6) is 41.1. After the interpolation of the BNLM and the RNN-BNLM models perplexity decreases with 11.3. This means that around 27\% of potential perplexity reduction can be recovered during the conversion of RNNLM to BNLM. If we were able to utilize RNN 4-gram (64.2) in the downstream task, this recovery rate could go up to around 52\%.

\subsubsection{Morph-based Models}

Just like in the case of word-based models, morph-based RNN language models significantly outperform BNLMs for every context size (see Table~\ref{morph_ppl}) as BNLM perplexities saturate at around 5 or 6-grams. Although morph-based perplexities are lower than word-based ones, note that the two perplexities can not be directly compared, since the vocabulary size of the two model types differs (50k vs. 30k).

\begin{table}[t]
\renewcommand{\tabcolsep}{1.7mm}
\centering
\caption{Perplexities of morph-based backoff n-gram, RNN n-gram and backoff approximated RNN language models as function of n-gram order}
\begin{tabular}{ccccccc}
\hline
\textbf{order} & \textbf{BNLM} & \textbf{\begin{tabular}[c]{@{}c@{}}RNN-\\BNLM\\ 100M\end{tabular}} & \textbf{\begin{tabular}[c]{@{}c@{}}BNLM + \\ RNN-BNLM\\ 100M\end{tabular}} & \textbf{\begin{tabular}[c]{@{}c@{}}RNN-\\BNLM\\ 1B\end{tabular}} & \textbf{\begin{tabular}[c]{@{}c@{}}BNLM + \\ RNN-BNLM\\ 1B\end{tabular}} & \textbf{\begin{tabular}[c]{@{}c@{}}RNN\\ n-gram\end{tabular}} \\ \hline
2              & 120.7         & 127.6                                                            & 114.9                                                                      & 122.6                                                          & 113.4                                                                    & 112.2                                                         \\
3              & 83.0          & 87.7                                                             & 74.1                                                                       & 80.2                                                           & 71.1                                                                     & 69.7                                                          \\
4              & 76.2          & 80.9                                                             & 66.6                                                                       & 71.8                                                           & 62.7                                                                     & 57.5                                                          \\
5              & 74.7          & 79.6                                                             & 64.9                                                                       & 70.1                                                           & 60.8                                                                     & 52.1                                                          \\
6              & 74.4          & 79.4                                                             & 64.5                                                                       & 69.8                                                           & 60.3                                                                     & 48.7                                                          \\
8              &               &                                                                  &                                                                            &                                                                &                                                                          & 45.7                                                          \\
10             &               &                                                                  &                                                                            &                                                                &                                                                          & 43.4                                                          \\
14             &               &                                                                  &                                                                            &                                                                &                                                                          & 43.2                                                          \\
18             &               &                                                                  &                                                                            &                                                                &                                                                          & 40.7                                                          \\ \hline
inf.           &               &                                                                  &                                                                            &                                                                &                                                                          & 40.2                                                          \\ \hline
\end{tabular}
\label{morph_ppl}
\end{table}

The morph-based results related to RNN-BNLM 100M model are also very similar to the word-based ones. The approximated model itself is a bit worse than the original BNLM; however, the interpolated model reduces perplexity with around 10-15\%. The question naturally arises: what if a much larger corpus is generated with the morph-based RNNLM. In order to answer this question we generated a ten times bigger corpus containing 1 billion morphs. As it can be seen in Table~\ref{morph_ppl} RNN-BNLM 1B significantly outperforms not just the RNN-BNLM 100M model but also the original BNLM. Moreover, the interpolated model (BNLM + RNN-BNLM 1B) further decreases perplexity, which suggests that in the future it may be useful to generate even larger corpora.

We calculated the perplexity recovery rate for the morph-based language models, as well. Interpolation of the morph-based BNLM and RNN-BNLM 100M models recover almost the same proportion of the potential perplexity reduction as the word-based models ({\raise.17ex\hbox{$\scriptstyle\sim$}}29\%). The 1 billion-morph-corpus, however, increases this rate to 40\%, which means almost half of the RNNLM-based perplexity improvement can be utilized in the ASR system. This way, morph-based BNLM approximations of RNNLMs got much closer to RNN 4-grams than in the case of word-based models. 

\subsection{Speech Recognition Experiments}

Perplexity is a useful measure to compare language models with a shared vocabulary. However, to assess the impact of different language modeling approaches on the ASR task, the best is to directly compare the automatic transcripts.

\subsubsection{Experimental Setup}

Classical hybrid Hidden Markov-Model (HMM) approach with Deep (feed-forward) Neural Network (DNN) probability distributions were used with three hidden layers consisting of 2500 neurons and output layer with 4907 neurons (senones). The acoustic model was trained on the 290 hours of the HCCSD 8 kHz sampled training data using the KALDI toolkit~\cite{Povey:192584}. As for acoustic features 13 dimensional MFCC (Mel-Frequency Cepstral Coefficients) were applied followed by LDA+MLLT~\cite{Povey:192584}. Shared-state context-dependent phone models were used, three states per phones. Acoustic and language model resources were compiled into weighted finite-state transducers and decoded with VoXerver~\cite{5999466} ASR decoder.

\subsubsection{Speech Recognition Results}

We performed single-pass decoding with the BNLM and RNN-BNLM models and calculated WER of each output (see Table~\ref{wer}). In order to ensure the fair comparison among the modeling approaches, we pruned each RNN-BNLM so that they had similar runtime memory footprint as the baseline BNLM models ({\raise.17ex\hbox{$\scriptstyle\sim$}}1GB). The interpolated language models (BNLM + RNN-BNLM) are also evaluated in a setup, where larger memory consumption is allowed.

\begin{table}[t]
\centering
\caption{Word Error Rate of the ASR system using the proposed language models}
\begin{tabular}{C{1.5cm}L{4cm}C{1.5cm}C{1.5cm}C{1.5cm}}
\hline
\textbf{\begin{tabular}[c]{@{}c@{}}Token\\ type\end{tabular}} & \textbf{Model}                      & \textbf{\begin{tabular}[c]{@{}c@{}}\# of\\ n-grams\\ {[}million{]}\end{tabular}} & \textbf{\begin{tabular}[c]{@{}c@{}}Memory\\ usage\\ {[}GB{]}\end{tabular}} & \textbf{\begin{tabular}[c]{@{}c@{}}WER\\ {[}\%{]}\end{tabular}} \\ \hline
\multirow{4}{*}{Word}                                         & BNLM                                & 5.0                                                                              & 1.3                                                                        & 29.2                                                            \\
                                                              & RNN-BNLM 100M                       & 4.8                                                                              & 0.9                                                                        & 30.2                                                            \\ \cline{2-5} 
                                                              & \multirow{2}{*}{BNLM+RNN-BNLM 100M} & 7.0                                                                              & 1.5                                                                        & 28.5                                                            \\
                                                              &                                     & 29.7                                                                             & 6.1                                                                        & 28.4                                                            \\ \hline
\multirow{7}{*}{Morph}                                        & BNLM                                & 5.1                                                                              & 1.0                                                                        & 28.7                                                            \\
                                                              & RNN-BNLM 100M                       & 8.5                                                                              & 1.1                                                                        & 28.9                                                            \\
                                                              & RNN-BNLM 1B                         & 7.2                                                                              & 0.9                                                                        & 28.6                                                            \\ \cline{2-5} 
                                                              & \multirow{2}{*}{BNLM+RNN-BNLM 100M} & 7.9                                                                              & 1.1                                                                        & 27.7                                                            \\
                                                              &                                     & 31.8                                                                             & 4.2                                                                        & 27.5                                                            \\ \cline{2-5} 
                                                              & \multirow{2}{*}{BNLM+RNN-BNLM 1B}   & 7.2                                                                              & 1.1                                                                        & 27.3                                                            \\
                                                              &                                     & 46.6                                                                             & 5.9                                                                        & 27.0                                                            \\ \hline
\end{tabular}
\label{wer}
\end{table}

ASR results of word-based language models show similar trends as perplexity results. The BNLM approximation of RNNLM (RNN-BNLM 100M) has a slightly higher WER than the baseline BNLM; however, the interpolated model (BNLM + RNN-BNLM 100M) outperforms both. The relative WER improvement of interpolated model compared to baseline BNLM is only around 2\%. Memory limit does not seem to have significant impact on the results. 

Replacing words with subwords in the baseline BNLM yields 2\% relative WER reduction, which is in accordance with our former findings~\cite{Tarjan2013b}. The morph-based BNLM trained on the 100-million-morph corpus (RNN-BNLM 100M) has larger WER than the original BNLM, just like in the case of word-based models. Using a ten times larger corpus to train the approximated model, however, seems to change the trend. Morph-based RNN-BNLM 1B model is the first approximated RNN model that outperforms a baseline BNLM by itself without interpolation. This observation underlines the importance of the size of the generated text. The difference between 100M and 1B models are also reflected in their interpolated counterparts. BNLM + RNN-BNLM 1B model can reduce WER of morph-based BNLM by 5\% or even 6\% if runtime memory consumption is not a restricting factor. 

All in all, the performance of morph-based BNLM approximations of RNN language models have exceeded our expectations. We managed to 
reduce the word error rate of our speech transcription system by 8\% relative by preserving real-time operation.

\section{Conclusions} \label{conclusions}

In this paper our aim was to improve our Hungarian conversational telephone speech recognition system by handling morphological richness of the language and transferring information from a recurrent neural language model to the back-off n-gram model used in the single-pass decoding. We compared various types of word-based and subword-based n-gram approximated RNNLMs and found that by generating a text with 1 billion morphs around 40\% of the perplexity improvement associated with the RNNLM can be transferred to the BNLM model. With the combination of subword modeling and RNNLM approximation, we were able to achieve 8\% relative WER reduction and preserve real-time operation of our conversational telephone speech recognition system. The perplexity we achieved with BNLM approximation of RNNLMs is roughly equal to the performance of an RNN 4-gram. The fact that RNN-BNLM was able to keep up with RNN n-gram until 4-gram is a quite promising result, but it also suggests that there is room for further improvement in utilizing higher order RNN n-grams in ASR decoding.

We consider the main contributions of our work are (1) presenting the first ASR results with using morph-based RNN-BNLMs in single-pass decoding; (2) comparing the performance of BNLMs and n-gram approximations of RNNLM (RNN n-gram models, RNN-BNLMs); (3) carrying out for the first time an evaluation of neural language models on very rich morphology Hungarian for speech recognition tasks on spontaneous speech; and (4) doing this preserving real-time operation capabilities by low delay.

In the future, we plan to place more emphasis on the study of OOV words. We would like to measure the recognition rate of OOV words and compare it among the word and morph-based language modeling approaches proposed in this paper. Moreover, we would like to evaluate models that extract features with character-based convolutional neural networks. Extending our work to other ASR tasks and share knowledge among them utilizing transfer learning methods is also a very promising direction of further research.

\section*{Acknowledgments}
The research was partly supported by the DANSPLAT (EUREKA\_15\_1\_2016-0019) project.

\bibliography{SLSP2019}
\bibliographystyle{splncs04}

\end{document}